%% file: example_paper.tex
\documentclass{article}

\usepackage{microtype}
\usepackage{graphicx}
\usepackage{booktabs} 
\usepackage{subcaption}
\usepackage{dsfont}
\usepackage{multirow}
\usepackage{amsmath}
\usepackage{amssymb}

\DeclareMathOperator*{\argmin}{argmin}  
\DeclareMathOperator*{\sign}{sign}  

\usepackage[draft]{hyperref}

\usepackage[table, dvipsnames]{xcolor}

\usepackage{tikz,array,calc}
\usetikzlibrary{decorations.pathreplacing}

\newcolumntype{C}[1]{>{\centering\arraybackslash}p{#1}}
\usetikzlibrary{arrows,automata,positioning}

\newcommand\tikzmark[2]{%
\tikz[remember picture,overlay] 
\node[inner sep=0pt,outer sep=2pt] (#1){#2};%
}
\newcommand\link[3]{%
\begin{tikzpicture}[remember picture, overlay]
\draw [decorate,decoration={brace,amplitude=5pt,raise=8pt}] 
(#1.north west)--(#2.north east) node[above=12pt,midway,text width=2cm,align=center]{#3};
\end{tikzpicture}%
}

\newcolumntype{C}[1]{>{\centering\arraybackslash\hspace{0pt}}m{#1}}

\newcommand\toplink[3]{%
\begin{tikzpicture}[remember picture, overlay]
\draw [decorate,decoration={brace,amplitude=5pt,raise=25pt}] 
(#1.north west)--(#2.north east) node[above=32pt,midway,text width=2cm,align=center]{#3};
\end{tikzpicture}%
}

\usepackage[accepted]{icml2020}

\icmltitlerunning{Correlated Feature Selection with Extended Exclusive Group Lasso}

\begin{document}

\twocolumn[
\icmltitle{Correlated Feature Selection with Extended Exclusive Group Lasso}



\icmlsetsymbol{equal}{*}

\begin{icmlauthorlist}
\icmlauthor{Yuxin Sun}{ucl_cs}
\icmlauthor{Benny Chain}{ucl_immu,ucl_cs}
\icmlauthor{Samuel Kaski}{fi_aal}
\icmlauthor{John Shawe-Taylor}{ucl_cs}
\end{icmlauthorlist}

\icmlaffiliation{ucl_cs}{Department of Computer Science, University College London, London, United Kingdom}
\icmlaffiliation{ucl_immu}{Division of Infection and Immunity, University College London, London, United Kingdom}
\icmlaffiliation{fi_aal}{Department of Computer Science, Aalto University, Espoo, Finland}

\icmlcorrespondingauthor{Yuxin Sun}{yuxin.sun.13@ucl.ac.uk}

\icmlkeywords{Feature selection, correlation, exclusive group Lasso, large-scale learning, convex optimization}

\vskip 0.3in
]



\printAffiliationsAndNotice{}  

\begin{abstract}
In many high dimensional classification or regression problems set in a biological context, the complete identification of the set of informative features is often as important as predictive accuracy, since this can provide mechanistic insight and conceptual understanding. Lasso and related algorithms have been widely used since their sparse solutions naturally identify a set of informative features. However, Lasso performs erratically when features are correlated. This limits the use of such algorithms in biological problems, where features such as genes often work together in pathways, leading to sets of highly correlated features. In this paper, we examine the performance of a Lasso derivative, the exclusive group Lasso, in this setting. We propose fast algorithms to solve the exclusive group Lasso, and introduce a solution to the case when the underlying group structure is unknown. The solution combines stability selection with random group allocation and introduction of artificial features. Experiments with both synthetic and real-world data highlight the advantages of this proposed methodology over Lasso in comprehensive selection of informative features.
\end{abstract}


\section{Introduction}
\label{sec: intro}

New technologies, especially the introduction of massively parallel Next Generation Sequencing of DNA have generated an explosion of very high dimensional biological and biomedical data sets. Intensively studied examples include global transcriptomics, typically involving the simultaneous measurement of 20--100 thousand RNA transcripts, and adaptive immune repertoires, typically involving millions of different RNA or DNA sequences per sample. Generally, these data sets are very sparse, in the sense that only a small proportion of the measurements (features) are relevant to the biological process being investigated, while the rest contribute only noise. Furthermore, features often show high levels of correlation, since they represent the activity of biological components acting together in groups to perform a task. Neither the correlation structure, nor the classification of features as informative or not, is usually known a priori. 

In this context, identifying the complete set of informative features connected to a particular biological process or outcome is an important task in itself, since it can provide fundamental mechanistic insights into how the process works, and is regulated, or misregulated in disease. Lasso~\cite{lasso} offers a promising approach since the sparse solution given by the $\ell_1$-norm regularization corresponds naturally to the set of informative features. However, Lasso has some limitations in optimal feature selection. First, Lasso finds at most $m$ features when sample size $m$ is much smaller than feature size $n$. This is often the case in many real-world problems. Secondly, choosing the regularization parameter is non-trivial, and optimizing classification does not always give the optimal solution in terms of accurate feature selection~\cite{lasso_pred_select, ns_two_step_lasso_dup}. Thirdly, Lasso performs erratically when features are correlated, selecting only a few features from each group of correlated features~\cite{elastic}. Theoretical analysis shows that for Lasso to select correct features, informative features (features with nonzero weights) cannot be overly correlated with irrelevant features (features with zero weights), or other informative features~\cite{ir_cond, pdw, lq_balls}. Therefore, Lasso will not discover all informative features when there is a strong correlation structure in the data.


A number of Lasso-type algorithms have therefore been developed (Figure~\ref{fig: compare norm}), which introduce a group structure into the global feature set. Group Lasso~\cite{group_lasso} forces inter-group sparsity through an $\ell_{2,1}$-norm regularization so that features from the same group, such as genes in the same biological pathways, are either selected or deselected simultaneously. However, group Lasso requires the group structure to be predefined through prior knowledge, while in many biological problems, such group structure is unknown. In contrast, the exclusive group Lasso~\cite{excl} adopts a complementary approach, introducing an $\ell_{1, 2}$-norm regularization term, which forces intra-group sparsity but relaxes inter-group sparsity. Thus only a few features within each group are selected. If co-correlated informative features are allocated to different groups, the exclusive group Lasso offers a potential solution to the problem of selecting correlated features.
\input{fig/reg.tex}

Although the properties of exclusive group Lasso have been investigated in several studies, these have focused either on examples where group structure was known, or where the informative features were uncorrelated, or where an proper group number was required~\cite{excl, excl_unsupervised, excl_kbc, excl_cone, excl_prob, excl_sparse_coding}. In this paper we examine the performance of exclusive group Lasso in the accurate selection of synthetic and real-world correlated features, and introduce new methods for relaxing its limitations.

In Section~\ref{sec: lasso}, we examine the limitations of Lasso on synthetic data and real-world immunological problems, which motivates the evolution of exclusive group Lasso in Section~\ref{sec: excl} to select biologically meaningful features. In this section, we develop efficient algorithms for solving the exclusive group Lasso. Lasso can be solved by efficient coordinate descent algorithms~\cite{lasso_coor_desc_org_dup, lasso_coor_desc_org, lasso_coor_desc}. In an analogous manner, group Lasso can be solved by fast implementation of block coordinate descent~\cite{group_lasso_coor_desc}. Other strategies such as active set have also been used in Lasso~\cite{lasso_coor_desc} and group Lasso~\cite{group_lasso_gene_dup, group_active}. In this paper we extend the algorithm proposed in~\cite{excl} in a different re-weighting approach and develop it to carry out coordinate descent with active set to achieve a fast solution for exclusive group Lasso. Section~\ref{sec: rand} proposes random group allocation with stability selection~\cite{stab_select} and artificial features for solving the problem of unknown group structure. Synthetic experiments that compare Lasso and our methodology are also included in Section~\ref{sec: rand}. Performance of the proposed methodology on real-world problems is validated in Section ~\ref{sec: cmv}. We summarize our main findings and contributions in Section~\ref{sec: concl}. 

\paragraph{Notation} Throughout this paper, we represent matrices as upper case, non-bold letters. Vectors are represented by lower-case, bold letters. We denote by $\mathbf{y} \in \mathds{R}^m$ the input labels and by $\mathbf{w} \in \mathds{R}^n$ the weight vector. We let $X$ represent an $m \times n$ matrix of input data, $X_j$ denote the $j$-th feature, and $\mathbf{x}_{(i)}$ denote the $i$-th sample. The term ``relevant features'' or ``informative features'' represents features with nonzero weights such that $\{X_j : \mathbf{w}_j = 0\}$, and ``irrelevant features'' represents features with zero weights where $\{X_j : \mathbf{w}_j = 0\}$. We assume the $X$ are standardized by default.



\section{Lasso and correlation}
\label{sec: lasso}

\begin{figure}[ht]
\begin{center}
\centerline{\includegraphics[width=\columnwidth]{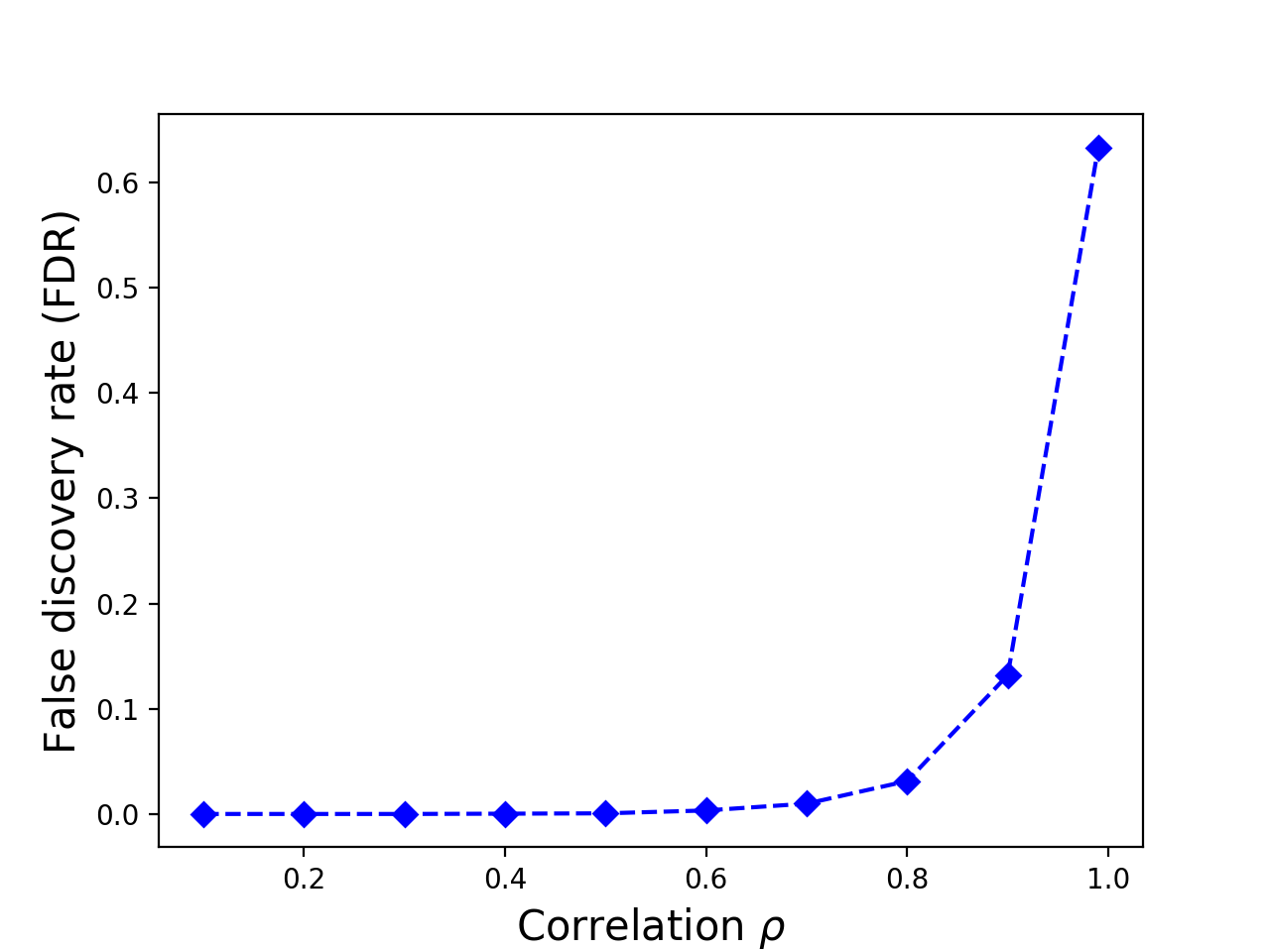}}
\caption{FDR in a feature selection task as a function of the strength of the correlation among the informative features. Selection threshold was set such that all informative features become chosen. Selection method: Lasso, on 200 random datasets.}
\label{fig: fdr}
\end{center}
\vskip -0.2in
\end{figure}

In Section~\ref{sec: intro}, we have reviewed that Lasso does not perform well on correlated features. Let us consider a simple experiment to verify this. We generate a dataset such that 
\begin{equation}
\mathbf{y} = X \mathbf{w} + \epsilon
\end{equation}
where $\epsilon_i \sim \mathcal{N}(0, \sigma^2), i = 1, \cdots, m$, is Gaussian noise, informative features $X_S \sim \mathcal{N}(0, \Sigma)$, and irrelevant features $X_{-S} \sim \mathcal{N}(0, 1)$. We split the first $n_S = 50$ features into two groups of 25. Features in the same group are pairwise correlated with correlation $\rho \in \{0.1, 0.2, \cdots, 0.9, 0.99\}$. We set the weight vector $\mathbf{w} = [s_1 w_S, s_2 w_S, \cdots, s_{n_S} w_S, 0, \cdots, 0], s_i \in \{-1, +1\}$, where the first $50$ entries equal to $w_S = 0.2$ with random signs. Therefore, the first $50$ features are relevant and informative, while the remaining features are irrelevant. Figure~\ref{fig: fdr} shows the false discovery rate (FDR) of Lasso with the largest possible $\lambda$ so that all informative features become discovered based on 200 random datasets. As $\rho$ increases, Lasso can only select all informative features at the cost of including irrelevant features.

To study the performance of Lasso on real-world biological problems, we consider a large dataset~\cite{cmv}. The study reported the T-cell receptors (TCRs) from a cohort of healthy individuals, a proportion of which are known  to be infected by human cytomegalovirus (CMV), a common  virus which does not cause any symptoms in otherwise healthy  individuals. A very small proportion of T cells are  known to recognize CMV, and control its growth. A fundamental dogma of the current model of adaptive immunity is the clonal selection theory~\cite{clonal_selection}, which proposes that T cells recognizing a virus  proliferate, and that CMV-related TCRs are therefore more likely to be detected in individuals infected with CMV. The task which the study therefore addressed was to identify a small set of TCRs which would be over-represented in the infected group (reflecting exposure to the virus) and could be used to predict which individuals were carrying the virus. The dataset contained two independent US cohorts, Cohort 1 and 2, for training and test. To further test the performance, we also use another geographically independent dataset (Cohort 3) that was collected by Belgium researchers~\cite{cmv_belgium}. Table~\ref{tab: cmv data} summarizes the three cohorts.

\begin{table}[tbp]
\caption{Description of the CMV dataset.}
\label{tab: cmv data}
\vskip 0.15in
\begin{center}
\begin{small}
\begin{sc}
\begin{tabular}{cccc}
\toprule
 &  Cohort 1 & Cohort 2 & Cohort 3 \\
 \midrule
\#CMV+ & 289 & 51 & 9 \\
\#CMV- & 352 & 69 & 24 \\
\midrule
\#TCR & 89,840,865 & 20,829,966 & 2,204,828 \\
\bottomrule
\end{tabular}
\end{sc}
\end{small}
\end{center}
\vskip -0.1in
\end{table}

We used Cohort 1 for training, and Cohorts 2 and 3 for test. To improve the efficiency, we preselected TCRs that occurred in at least 10 samples in Cohort 1. This reduced the number of unique TCRs to 241,008. We applied a 10-fold cross-validation for parameter tuning. As 35 samples were published for 33 patients in Cohort 3, we also included the two additional measurements without affecting the training model. We implemented a two stage strategy, identifying informative TCRs using $\ell_1$-regularized SVM (the classification version of Lasso) followed by RBF or linear $\ell_2$-regularized SVM for classification. Classification results are shown in Table~\ref{tab: cmv acc lasso}. Among the 1950 TCRs selected by the $\ell_1$-SVM as informative features, only 44 were found to be over-represented in individuals infected with CMV. 



\begin{table}[tb!]
\centering
\caption{Classification accuracy on CMV dataset.}
\label{tab: cmv acc lasso}
\vskip 0.15in
\begin{center}
\begin{small}
\begin{sc}
\begin{tabular}{lcc}
\toprule
  &  Linear-SVM &  RBF-SVM \\   
  \midrule
Cohort 2 & 0.65  & 0.65  \\  
Cohort 3 & 0.51  & 0.63  \\  
\bottomrule
\end{tabular}
\end{sc}
\end{small}
\end{center}
\vskip -0.1in
\end{table}

Clearly, the $\ell_1$-regularized algorithm neither selected the biologically meaningful features, nor achieved high accuracy with the selected features. In the meantime, it is not possible to apply group Lasso to this problem as underlying group structure is unknown. This motivates us to apply and adjust exclusive group Lasso to select correlated CMV-related TCRs without knowing the explicit group structure.


\section{From Lasso to exclusive group Lasso}
\label{sec: excl}


The exclusive group Lasso~\cite{excl} is defined as 
\begin{equation}
\label{eq: l12}
\min_\mathbf{w} f(\mathbf{w}) + \lambda \sum_{g \in \mathcal{G}} ||\mathbf{w}_g||_1^2
\end{equation}
where $ f(\mathbf{w})$ is a convex loss function, $\mathcal{G}$ represents group allocation and $g$ represents indices of group $g \in \mathcal{G}$. \citet{excl} presented an iterative algorithm to solve the optimization problem. Let us denote by $I_g^i \in \{0, 1\}$ a binary group indicator of the $i$-th feature. The algorithm proceeds by alternating the following steps.
\begin{align}
&\text{Solve $\mathbf{w}$:} && \nabla_\mathbf{w} f(\mathbf{w}) + 2 \lambda F \mathbf{w} = 0     \label{eq: update w}\\
&\text{Compute $F$:} &&  F_{ii} = \left(   \sum_{g \in \mathcal {G}} \frac{   I_g^i  ||\mathbf{w}_g||_1}{|\mathbf{w}_i|}     \right), F_{ij} = 0, i \neq j    \label{eq: update f}
\end{align}
\subsection{A re-weighting scheme}

In this section, we provide an alternative solution to the exclusive group Lasso particularly by solving $\mathbf{w}$ in \eqref{eq: update w} via a re-weighting approach. For convenience we rewrite \eqref{eq: l12} as 
\begin{equation}
\label{eq: l12 tran}
\min_\mathbf{w} f(X \mathbf{w} , \mathbf{y} ) + \lambda \sum_g ||\mathbf{w}_g||_1^2
\end{equation}
which can be further rewritten as
\begin{equation}
\label{eq: tran}
\min_\mathbf{w} f(X \mathbf{w},  \mathbf{y} ) + \lambda \mathbf{w}^T F \mathbf{w}.
\end{equation}
Let $\tilde{ \mathbf{w} } = \sqrt{F} \mathbf{w}$ and substitute $\mathbf{w}$ with $\sqrt{F}^{-1} \tilde{ \mathbf{w} }$; the loss function then becomes $f(X \sqrt{F}^{-1} \tilde{ \mathbf{w} },  \mathbf{y} )$. By denoting $X \sqrt{F}^{-1}$ with $\tilde{X}$, \eqref{eq: tran} can be rewritten as
\begin{equation}
\label{eq: tran final}
\min_\mathbf{w} f(\tilde{X} \tilde{ \mathbf{w} },  \mathbf{y} ) + \lambda \tilde{ \mathbf{w} }^T \tilde{ \mathbf{w} }.
\end{equation}
Therefore, the exclusive group Lasso converts into an optimization problem with re-weighted features $\tilde{X}$ and weights $\tilde{ \mathbf{w} }$, which can make use of the widely-studied $\ell_2$-regularized optimization techniques. An additional benefit of the re-weighting step is that it allows us to introduce additional constraints on the values of the weights. For example, biological knowledge would suggest that CMV infection would lead to over-representation of some TCRs, but not necessarily to under-representation of TCRs in the infected group. To capture this biological prior knowledge, we imposed the constraint that weights should be all be of the same sign. We summarize the re-weighting approach with adjustments in Algorithm~\ref{alg: trans pos weight}.

\begin{algorithm}[tb!]
   \caption{Exclusive group Lasso with re-weighting}  
   \label{alg: trans pos weight}
\begin{algorithmic}

\STATE {\bfseries Input:} feature space $X$, labels $ \mathbf{y} $, regularization parameter $\lambda$, group indicator matrix $I$, groups $\mathcal{G}$, indicator of constraints for biologically meaningful features $r \in \{0, 1\}$, constrained class $c \in \{-1, +1\}$\\
   \STATE $\mathbf{w} \gets \frac{1}{n}$\\
   \WHILE{not converged}
   \IF {$ r = 1$}
   	\STATE{$\mathbf{w}_{i: \sign(\mathbf{w}_i) = c} \gets \epsilon$, where $\epsilon$ is a very small number}\\
   \ENDIF
   \STATE{$F_{ii} \gets \frac{1}{|\mathbf{w}_i|} \sum_{g \in \mathcal{G}} I_g^i ||\mathbf{w}_g||_1$}\\
   \STATE{$\tilde{X} \gets X \sqrt{F}^{-1}$}\\
   \STATE{$\tilde{ \mathbf{w} }  \gets \argmin_{\tilde{ \mathbf{w} }}  f(\tilde{X} \tilde{ \mathbf{w} },  \mathbf{y} ) + \lambda \tilde{ \mathbf{w} }^T \tilde{ \mathbf{w} }$}\\
   \STATE{$\mathbf{w} \gets \sqrt{F}^{-1} \tilde{ \mathbf{w} }$}
   \ENDWHILE
   \STATE \textbf{return} $\mathbf{w}$
   
   \end{algorithmic}
\end{algorithm}

\subsection{An iterative approach}

Exclusive group Lasso with square loss is given by
\begin{equation}
\label{eq: l12 sq loss}
\min_\mathbf{w} ||\mathbf{y} - X \mathbf{w}||^2 + \lambda \sum_{g \in \mathcal{G}} ||\mathbf{w}_g||_1^2
\end{equation}
\citet{coor_desc_converge, block_coor_desc_converge} proved that coordinate descent algorithms converge for the problem
\begin{equation}
\label{eq: coor desc general}
\min_x g(x) + \sum_i h_i (x_i)
\end{equation}
under these conditions: 1) $g(x)$ is convex and differentiable; 2) each $h_i(x_i)$ is convex; and 3) when $x_i$ is a vector, it cannot overlap with other vectors. Let $-g$ represent the complement of $g$ in the set of features. Then \eqref{eq: l12 sq loss} can be solved by iteratively solving subproblems of individual blocks,
\begin{equation}
\label{eq: l12 sub}
\min_{\mathbf{w}_g} ||\mathbf{y}_g - X_g \mathbf{w}_g||^2 + \lambda ||\mathbf{w}_g||_1^2,
\end{equation}
where
\begin{equation}
\mathbf{y}_g = \mathbf{y} - X_{-g} \mathbf{w}_{-g}.
\end{equation}
Solving problem~\eqref{eq: l12 sub} is equivalent to solving
\begin{equation}
\label{eq: l12 sub z}
\begin{aligned}
&\min_{\mathbf{w}_g} ||\mathbf{y}_g - X_g \mathbf{w}_g||^2 + \lambda z ||\mathbf{w}_g||_1\\
& s.t.  \quad z = ||\mathbf{w}_g||_1
\end{aligned}
\end{equation}
which can be further rewritten as 
\begin{equation}
\label{eq: l12 sub lambda prime}
\min_{\mathbf{w}_g} ||\mathbf{y}_g - X_g \mathbf{w}_g||^2 + \lambda' ||\mathbf{w}_g||_1^2.
\end{equation}
Now \eqref{eq: l12 sub} has become converted to Lasso with a new regularization parameter $\lambda' = \lambda z$. Theoretically we can solve \eqref{eq: l12 sub lambda prime} with every possible $\lambda'$ using Lasso, and the optimal solution is recovered when the solution of \eqref{eq: l12 sub lambda prime} satisfies $||\mathbf{w}^*_g||_1  =  \frac{\lambda'}{ \lambda}$, under the current choice of $\lambda$. In practice, solving the problem is fairly fast with bisection approaches, given the piecewise linear property of Lasso~\cite{lasso_path}. 


Algorithm~\ref{alg: bisection} shows a bisection algorithm to find the optimal solution to \eqref{eq: l12 sub z} and Algorithm~\ref{alg: iter} shows solution to \eqref{eq: l12 sq loss} with block coordinate descent. It is possible to extend Algorithm~\ref{alg: iter} to other types of loss functions, as long as the loss function is convex and differentiable and the corresponding $\ell_1$-norm solver is piecewise linear.

\begin{algorithm}[tb]
   \caption{Bisection algorithm to solve $\mathbf{w}_g$}  
   \label{alg: bisection}
\begin{algorithmic}
   \STATE {\bfseries Input:} features of current group $X_g$, labels $\mathbf{y}_g$, regularization parameter $\lambda$, lower bound $\lambda_1$, upper bound $\lambda_2$\\
   \STATE $\lambda' \gets \frac{(\lambda_1 + \lambda_2)}{2} $\\
   \STATE $\mathbf{w}_g \gets \text{Lasso}(X_g, \mathbf{y}_g,    \lambda')$\\
   \IF{converged}
        \STATE {\textbf{return} $\mathbf{w}_g, \lambda'$}\\
    \ELSIF{$\lambda' <  \lambda ||\mathbf{w}_g||_1$}
        \STATE \textbf{return}  Bisection($X_g, \mathbf{y}_g, \lambda, \lambda_1, \lambda'$)\\
    \ELSE
        \STATE \textbf{return} Bisection($X_g, \mathbf{y}_g, \lambda, \lambda', \lambda_2$)\\
      \ENDIF   
   
\end{algorithmic}
\end{algorithm}

\begin{algorithm}[tb]
   \caption{Iterative exclusive group Lasso}  
   \label{alg: iter}
\begin{algorithmic}
  \STATE {\bfseries Input:} feature space $X$, labels $\mathbf{y}$, regularization parameter $\lambda$, lower bound $\lambda_1$, upper bound $\lambda_2$, groups $\mathcal{G}$\\
   \STATE $\mathbf{w} \gets \mathbf{0}$\\
   \WHILE{not converged}
   \FORALL{$ g \in \mathcal{G}$}
   \STATE{$\mathbf{y}_g \gets \mathbf{y} - X_{-g} \mathbf{w}_{-g}$}\\
   \STATE{$\mathbf{w}_g \gets \text{Bisection}(X_g, \mathbf{y}_g, \lambda, \lambda_1, \lambda_2)$}
   \ENDFOR
   \ENDWHILE
   \STATE \textbf{return} $\mathbf{w}$
        
 \end{algorithmic}
\end{algorithm}

\subsection{A faster solution with the active set of features}

 For large-scale datasets coordinate descent in $\ell_1$-regularized methods such as Lasso is often faster than with $\ell_2$-regularized methods like ridge regression. We can therefore reduce the computational time of Algorithm~\ref{alg: trans pos weight} by taking advantage of the sparsity of $\ell_1$-regularized algorithms such as Algorithm~\ref{alg: iter}. We first run a single iteration over all groups in Algorithm~\ref{alg: iter}, then apply Algorithm~\ref{alg: trans pos weight} with a reduced feature space of features receiving nonzero weights from the other algorithm until convergence. We usually need another cycle of Algorithm~\ref{alg: iter} on the complete feature space or to check KKT conditions to ensure convergence---if not converged, we simply run Algorithm~\ref{alg: trans pos weight} again on nonzero features. The double-iteration procedure is repeated until convergence. We summarize the combined algorithm in Algorithm~\ref{alg: active}.

\begin{algorithm}[tb]
   \caption{Exclusive group Lasso with active set}  
   \label{alg: active}
\begin{algorithmic}
\STATE {\bfseries Input:} feature space $X$, labels $ \mathbf{y} $, regularization parameter $\lambda$, lower bound $\lambda_1$, upper bound $\lambda_2$, group indicator matrix $I$, groups $\mathcal{G}$\\
   \STATE $\mathbf{w} \gets \mathbf{0}$\\
   \WHILE{not converged overall}
   \STATE{Run a single iteration over all groups of Algorithm~\ref{alg: iter}}\\
   \STATE{Run Algorithm~\ref{alg: trans pos weight} on features with nonzero weights until Algorithm~\ref{alg: trans pos weight} converges}\\
   \ENDWHILE
   \STATE \textbf{return} $\mathbf{w}$
   
   \end{algorithmic}
\end{algorithm}

Optimization in~\cite{excl, excl_prob} computed $\mathbf{w} = (X X^T + \lambda F )^{-1} X \mathbf{y}$ for \eqref{eq: update w}, which resulted in 10,221s even when $m = 1000, n = 5000, n_S = 100$. Therefore we compared the computational time of exclusive group lasso with active set (EGL-AS) and re-weighting scheme (EGL-RW). We randomly generated 1,000 or 10,000 samples with various numbers of features ($n$) and pairwise correlated informative features ($n_S$) at $\rho=0.9$. We set $|\mathbf{w}_S| = 0.3$ with random signs. Exclusive group Lasso was run with fixed groups on 10 $\lambda$s using Scikit-learn~\cite{sklearn} in Python 2.7. The computational times are reported in Table~\ref{tab: time}. There is a pronounced speed-up in convergence time once $n$ exceeds 5000. 

\begin{table}[tbp]
\caption{Computational time (s) of exclusive group Lasso.}
\label{tab: time}
\vskip 0.15in
\begin{center}
\begin{small}
\begin{sc}
\begin{tabular}{ccccc}
\toprule

$m$ & $n$ & $n_S$  & EGL-AS  & EGL-RW \\
 \midrule
1000 & 5000  & $ 100$ & 44.64 & 187.00 \\
1000 & 5000 & $500$ & 224.99 & 167.10 \\
1000 & 5000 & $1000$ &  511.57 &  190.25\\
\midrule
1000 & 10000 & $100$ & 53.08 & 564.58 \\
1000 & 10000 & $500$ & 244.06 & 574.34 \\
1000 & 10000 & $ 1000$ & 579.52 & 600.42 \\
\midrule
1000 & 20000  & $100$ & 82.13 & 2000.66 \\
\midrule
1000 & 50000  & $100$ & 980.87 & 14206.60 \\
\midrule
 10000 & 20000 & 100 & 1429.97 & 31332.65\\
 \midrule
 10000 & 50000 & 100 & 3947.49 & 173769.50\\
\bottomrule
\end{tabular}
\end{sc}
\end{small}
\end{center}
\vskip -0.1in
\end{table}


\section{Random group allocation and beyond}
\label{sec: rand}

\subsection{Fixed and random group allocation}

\begin{figure}[tbp]
\vskip 0.2in
\begin{center}
\centerline{\includegraphics[width=\columnwidth]{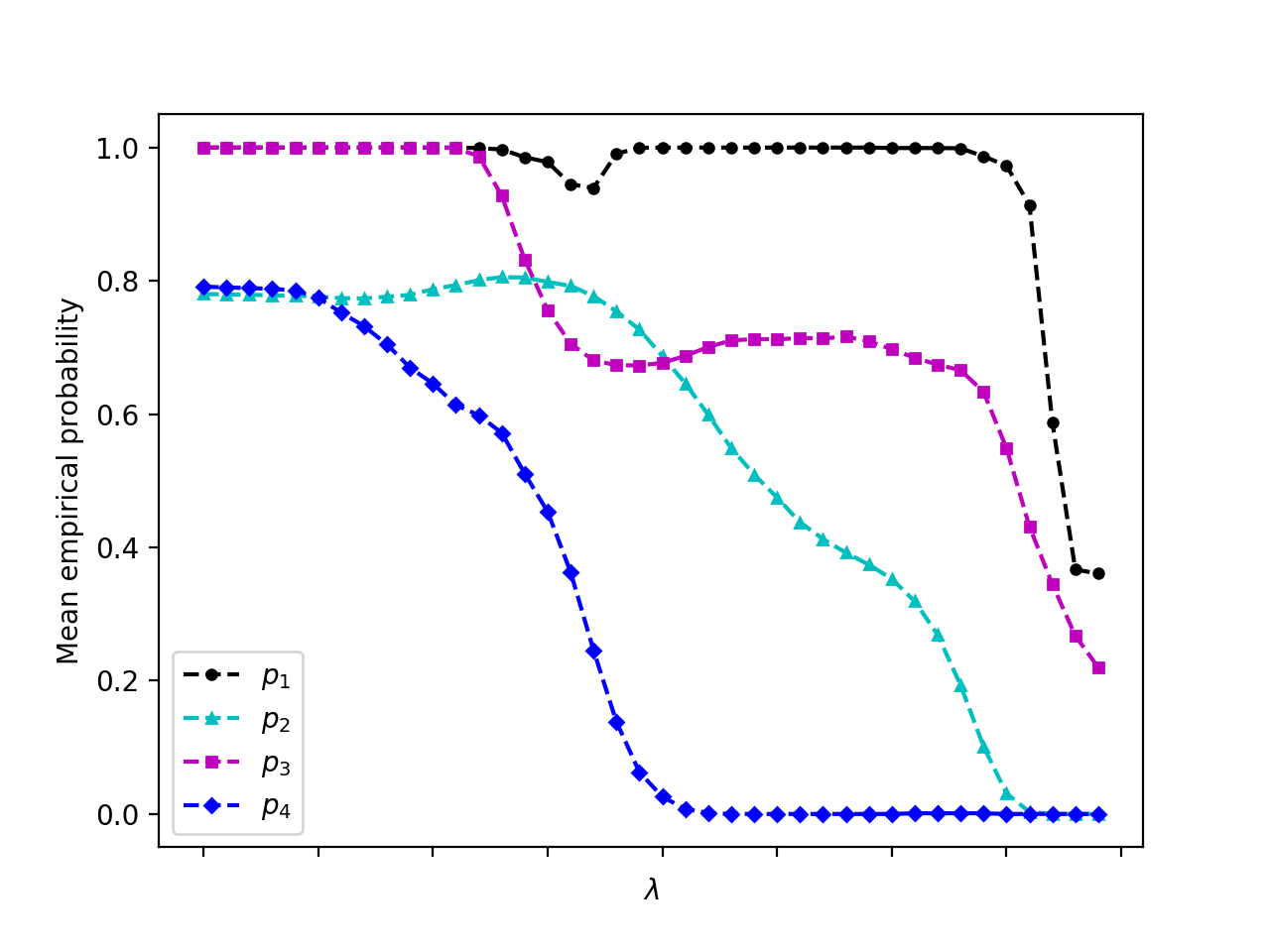}}
\caption{Empirical probabilities as a function of strength of regularization (regularization parameter $\lambda$). $p_1$: Prob[a relevant feature is selected $|$ no other relevant features exist in the same group]; $p_2$: Prob[an irrelevant feature is selected $|$ no relevant features exist in the same group]; $p_3$: Prob[a relevant feature is selected $|$ other relevant features exist in the same group]; $p_4$: Prob[an irrelevant feature is selected $|$ at least one relevant feature exists in the same group].}
\label{fig: prob}
\end{center}
\vskip -0.2in
\end{figure}

As group Lasso, exclusive group Lasso also requires reliable group allocation. Figure~\ref{fig: prob} shows the empirical probabilities that a feature is selected when other informative or irrelevant features coexist in the same group. The probabilities were averaged over 100 random datasets, each containing 50 relevant features and 150 irrelevant features. As expected, informative features are more likely to be discovered when no other informative features coexist in the same group, while the probability of selecting an irrelevant feature is lower if a group contains at least one informative feature. The ideal group allocation is therefore to allocate each informative (and correlated) feature into a separate group, and set the number of groups to the number of informative features. We refer to this ideal group allocation as ``fixed groups''. In most real-world cases informative features are unknown, so groups are allocated randomly (``random groups''). 

We compared the performance of Lasso and exclusive group Lasso with fixed and random groups on synthetic datasets that were generated such that $\mathbf{x}_{(i)} \sim \mathcal{N}(0, \Sigma), \mathbf{\epsilon}_i \sim \mathcal{N}(0, \sigma^2), i = 1, \cdots, m$, with $\Sigma$ being the covariance matrix. We inherit the weight vector in Section~\ref{sec: lasso} with the first $n_S$ entries being some scalar $w_S$ with random signs, and the last $n - n_S$ entries being 0. Noise followed a Gaussian distribution $\mathbf{\epsilon}_i \sim \mathcal{N}(0, \sigma^2)$. We computed $\mathbf{y} = X \mathbf{w} + \mathbf{\epsilon}$. To validate the performance on different covariances, four types of $\Sigma$ were considered.

\begin{enumerate}
\item \label{enu: useful corr} Example 1 (pairwise correlation): relevant features are pairwise correlated at correlation $\rho$. This simulates the case when active biological patterns such as genes and TCRs are correlated because of functional similarity.
\item \label{enu: rand pairwise} Example 2 (random pairwise correlation): relevant features are pairwise correlated at $\rho$, while some irrelevant features are correlated with randomly selected relevant features at correlation $\frac{\rho}{2}$. Similar real-world examples include discovery of virus-related patterns, which may share lower correlation with other virus mutations, e.g. a certain type of flu virus and its mutations. 
\item \label{enu: small block} Example 3 (blocked diagonal correlation): relevant features are grouped into equal-sized groups and features in the same group are pairwise correlated at $\rho$. If features from the same group are adjacent to each other, all entries in $\Sigma$ are zero, except for the blocks along the diagonal. This resembles the situation when desired features have underlying subgroup structure such as disease subtypes and active areas in brain scan images.
\item \label{enu: er graph} Example 4 (Erdos-Renyi correlation): relevant features are correlated at $\rho$ based on the connectivity of an Erdos-Renyi graph. This is a more complexed case of Example~\ref{enu: small block}, perhaps closer to real-world scenarios.
\end{enumerate}

We set $m = 100, n = 100, n_S = 30$. For Example~\ref{enu: rand pairwise}, $\rho$ was set to 0.6 because it was the maximum correlation under this setting. We used the same $\rho$ for Examples~\ref{enu: useful corr} and \ref{enu: small block}. In addition, 20 irrelevant features correlated with two randomly selected relevant features at correlation 0.3.  For Example~\ref{enu:  small block}, we used 5 blocks. For Example~\ref{enu: er graph}, $\rho = 0.3$ because 0.3 was the maximum value to ensure $\Sigma$ to be positive definite. The connection probability was set to $\frac{5}{29}$ to mimic the structure of Example~\ref{enu: small block}. We set $\sigma^2 = 1$ and $w_S = \{0.1, 0.3, 0.5\}$. The experiments were based on 50 randomly generated datasets. For random exclusive group Lasso, 50 random groups were used. We evaluated the feature selection results with F measure, a common metric in information retrieval. F measure is computed by
\begin{equation}
2 * \frac{precision * recall} {precision + recall}
\end{equation}
where $precision = \frac{n_S^*}{n^*}, recall = \frac{n_S^*}{n_S}$, $n_S$ is the number of relevant features, $n_S^*$ is the number of selected relevant features, and $n^*$ is the number of all selected features. Table~\ref{tab: naive} compares the results of Lasso and exclusive Lasso with fixed and random groups.

\begin{table}[t]
\caption{F measure of Lasso, exclusive group Lasso with fixed groups (EGL-F) and random groups (EGL-R). EGL outperformed Lasso in almost all experiments.}
\label{tab: naive}
\vskip 0.15in
\begin{center}
\begin{small}
\begin{sc}
\begin{tabular}{lcccc}
\toprule
 & $w_S$ & Lasso & EGL-F & EGL-R \\
 \midrule
 \multirow{ 3}{*}{Example 1} & 0.1 & 0.44 & 0.88 & 0.88 \\
 & 0.3 & 0.54 & 0.97 & 0.94 \\
 & 0.5 & 0.60 & 0.96 & 0.99 \\
  \midrule
\multirow{ 3}{*}{Example 2} &  0.1 & 0.41  & 0.86 & 0.70 \\
 & 0.3 & 0.51  & 0.98 & 0.93 \\
 & 0.5 & 0.56  & 0.97 & 0.91 \\
  \midrule
\multirow{ 3}{*}{Example 3} & 0.1 & 0.44  & 0.58 & 0.48 \\
 & 0.3 & 0.54  & 0.73 & 0.55 \\
 & 0.5 & 0.62  & 0.76 & 0.65 \\
  \midrule
\multirow{ 3}{*}{Example 4} & 0.1 & 0.47 & 0.44 & 0.46 \\
 & 0.3 & 0.52  & 0.58 & 0.53 \\
 & 0.5 & 0.56  & 0.61 & 0.54 \\

\bottomrule
\end{tabular}
\end{sc}
\end{small}
\end{center}
\vskip -0.1in
\end{table}

In most experiments, exclusive group Lasso with fixed groups outperformed Lasso. Exclusive group Lasso with random group allocation also outperformed Lasso, but performed worse than using fixed groups. However, fixed group allocation is usually not possible in real-world scenarios where group structure is unknown for most problems. We therefore introduce stability selection to improve the stability of feature selection with random group allocation.

\subsection{Stability selection on random groups}


Stability selection combines subsampling with high-dimensional selection algorithms. \citet{stab_select} showed that stability selection performs well even when Lasso fails to select correct features. We adapt stability selection with random group allocation in exclusive group Lasso in Algorithm~\ref{alg: stab select}. In practice, a threshold for selection probability is required to distinguish informative and irrelevant features. This can be done with cross-validation with an extra parameter to tune. In our experiments, we rely on $k$-means to partition probabilities for informative and irrelevant features. We applied stability selection with 50 iterations to the same datasets used in Table~\ref{tab: naive} and compared Lasso and exclusive group Lasso with fixed and random groups in Table~\ref{tab: stab select}. The introduction of stability selection provided moderate improvement of exclusive group Lasso with random groups in almost all cases. So far, we have tested the performance when $m = n$ under various correlation types in the above experiments. We further examine the results when $m < n$ where $m = 300, n = 1500, n_S = 30$, and $w_S = 0.5$. Table~\ref{tab: stab select large n} shows the average F measure using stability selection. Exclusive group Lasso with random groups substantially outperformed Lasso in all types of data under these high-dimensional conditions.  

\begin{algorithm}[tb]
   \caption{Stability selection}  
   \label{alg: stab select}
\begin{algorithmic}
   \STATE  {\bfseries Input:} input data $X$, labels $\mathbf{y}$, regularization parameter $\lambda$, number of iterations $N$, group $\mathcal{G}$\\
   \FOR {$ i = 1 : N$}
   \STATE {Subsample from $X$ without replacement to generate subsets $X_{I_i}, \mathbf{y}_{I_i}$, where $I_i \subseteq \{1, \cdots, m\},  |I_i| = \frac{m}{2}$ }\\
   \STATE {Run exclusive group Lasso with $X_{I_i}$, $\mathbf{y}_{I_i}$ to select features $\hat{S}_i^\lambda = \{j \mid \mathbf{w}_j^* \neq 0\}$}\\
   \STATE {Shuffle $\mathcal{G}$ if using random group allocation}
   \ENDFOR
   \STATE {$\Pi_j^\lambda \gets \frac{1}{N} \sum_{i=1}^{N} \mathds{I} \{ j \in \hat{S}_i^\lambda \}  $}\\
   \STATE {$\hat{S}^{stable} \gets \{j: \hat{\Pi} _j^\lambda \geq \pi \}$}
\end{algorithmic}
\end{algorithm}

\begin{table}[tbp]
\caption{F measure of Lasso, EGL-F, and EGL-R with stability selection (S). (S) showed improvements and EGL-R(S) performed better than EGL-R particularly when EGL did not perform well.}
\label{tab: stab select}
\vskip 0.15in
\begin{center}
\begin{small}
\begin{sc}
\begin{tabular}{lcccc}
\toprule
 & $w_S$ & Lasso(S)  & EGL-F(S) & EGL-R(S) \\
 \midrule
 \multirow{3}{*}{Example 1} & 0.1 & 0.47  & 0.90 & 0.84 \\
 & 0.3 & 0.56  & 0.97 & 0.95 \\
 & 0.5 & 0.61  & 0.96 & 0.94 \\
 \midrule
\multirow{ 3}{*}{Example 2} & 0.1 & 0.47  & 0.88 & 0.82 \\
 & 0.3 & 0.54  & 0.99 & 0.97 \\
 & 0.5 & 0.58  & 0.97 & 0.95 \\
 \midrule
\multirow{ 3}{*}{Example 3} & 0.1 & 0.47  & 0.58 & 0.54 \\
 & 0.3 & 0.56  & 0.77 & 0.68 \\
 & 0.5 & 0.56  & 0.77 & 0.68 \\
 \midrule
\multirow{ 3}{*}{Example 4} & 0.1 & 0.48  & 0.51 & 0.49 \\
 & 0.3 & 0.66  & 0.68 & 0.62 \\
 & 0.5 & 0.77  & 0.74 & 0.64 \\
\bottomrule
\end{tabular}
\end{sc}
\end{small}
\end{center}
\vskip -0.1in
\end{table}


\begin{table}[tbp]
\centering
\caption{F measure for $n = 1500, w_S=0.5$. Compared to $n=100$, Lasso performed worse while EGL had similar F measure.}
\label{tab: stab select large n}
\vskip 0.15in
\begin{center}
\begin{small}
\begin{sc}
\begin{tabular}{lcccc}
\toprule
 & Lasso(S)  & EGL-F(S) & EGL-R(S) \\
 \midrule
  Example 1 & 0.23  & 0.99 & 0.98 \\
 Example 2 & 0.24  & 0.96 & 0.95 \\
 Example 3 & 0.28  & 0.69 & 0.64 \\
 Example 4 & 0.42  & 0.72 & 0.59 \\
 \bottomrule
 \end{tabular}
 \end{sc}
\end{small}
\end{center}
\vskip -0.1in
 \end{table}


\subsection{Artificial features}

Recalling Figure~\ref{fig: prob}, it is clear that when a group contains only irrelevant features, some irrelevant features from this group are selected. However, to ensure that two informative features are rarely found in the same group, especially taking into consideration that the real number of informative features is usually unknown, the number of groups will be chosen as larger than the estimated number of informative features. A significant proportion of the groups may therefore contain no informative feature. Hence, we propose a novel method to further reduce the risk of including too many irrelevant features by inserting some artificially generated informative features into feature groups.

For a regression problem, this approach involves:

\begin{enumerate}
\item Randomly generate $X' \in R^{m \times n_S}, \mathbf{w}' \in R^ {n_S}$ and new labels $\mathbf{y}' = \mathbf{y} + X' \mathbf{w}'$;
\item Insert a feature from $X'$ into each group;
\item Find solution with $[X, X']$ and $\mathbf{y}'$;
\item Remove selected artificial features after convergence.
\end{enumerate}

The procedure for classification is slightly different since the labels need to remain unchanged. We let the artificial features corresponding to each class follow different distributions, so the first step of the above method becomes


\begin{enumerate}
\item Randomly generate $X' \in R^{m \times n_S}$, where rows of positive samples $\mathbf{x}'_{(+)} \sim \mathcal{N}(\mathbf{\mu}_1, \Sigma_1)$, and rows of negative samples $\mathbf{x}'_{(-)} \sim \mathcal{N}(\mathbf{\mu}_2, \Sigma_2)$. $\Sigma_1, \Sigma_2$ are diagonal covariance matrices.
\end{enumerate}

In practice, we find Gaussian distributed $X'$ and $\mathbf{w}'$, such as $X'_{ij} \sim \mathcal{N}(\mu_1, \sigma_1^2), \mathbf{w}'_j \sim \mathcal{N}(\mu_2, \sigma_2^2)$, generally achieve good performance. $\mathbf{w}'$ of fixed values also works well in some experiments.

\begin{figure}[tb!]
\vskip 0.2in
    \begin{subfigure}[h]{\linewidth}
        \centering
        \makebox[\textwidth][c]{\includegraphics[width = \columnwidth]{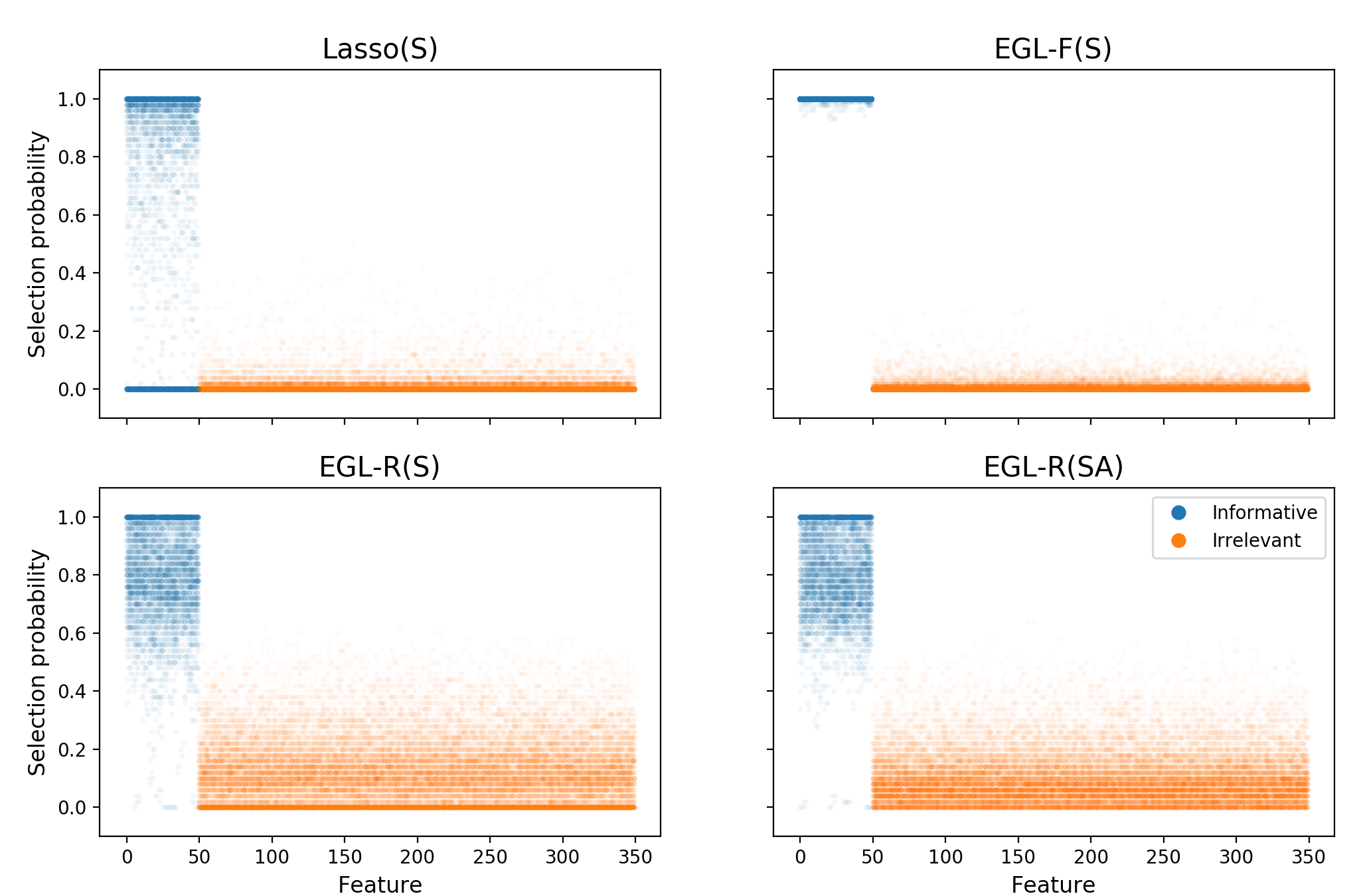}}
        \caption{Example~\ref{enu: small block}}
    \end{subfigure}
    \begin{subfigure}[h]{\linewidth}
        \makebox[\textwidth][c]{\includegraphics[width = \columnwidth]{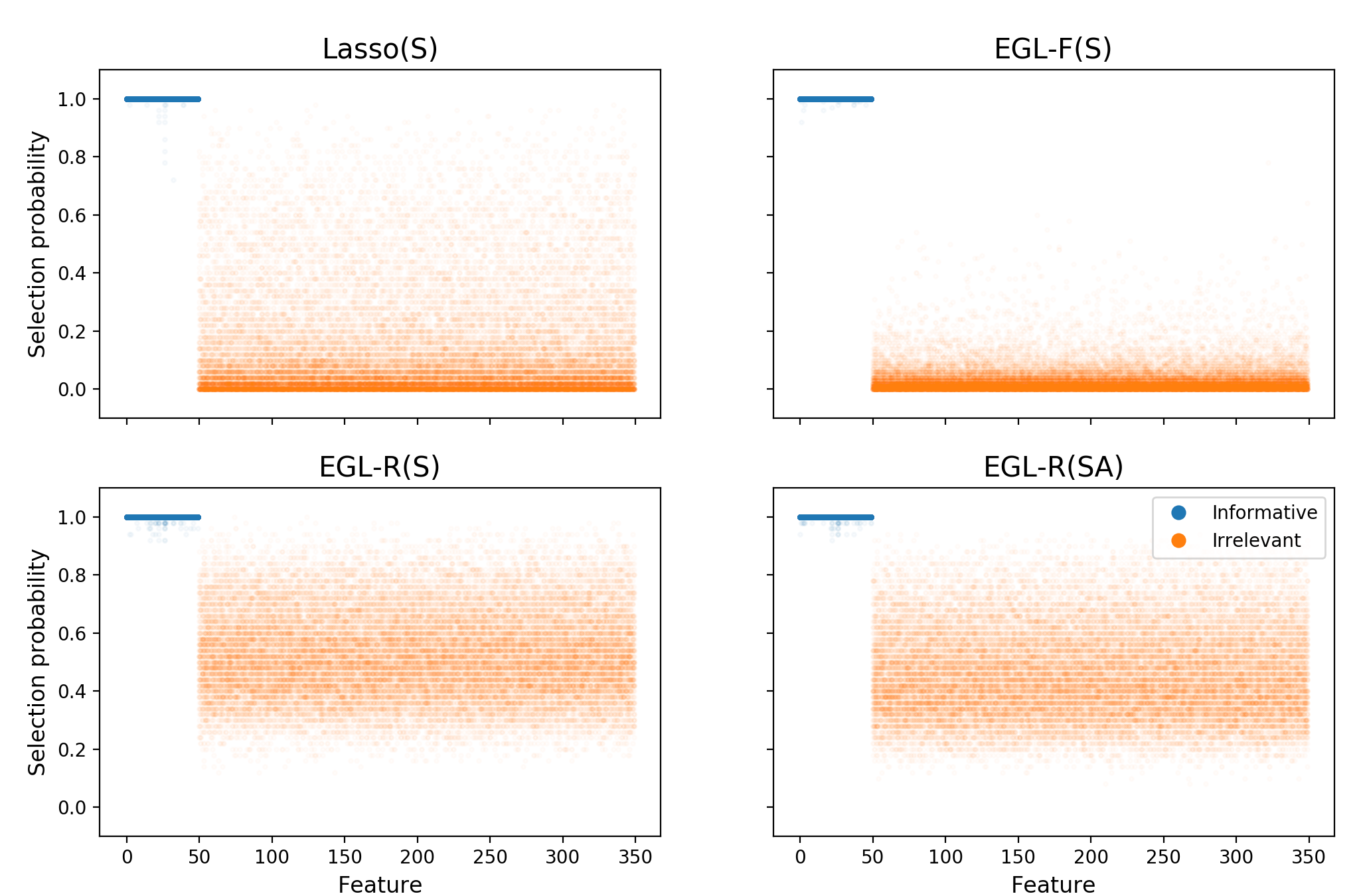}}
        \caption{Example~\ref{enu: er graph}}
    \end{subfigure}
\caption{``Stacked'' selection probabilities from 100 random datasets. EGL clearly distinguished informative and irrelevant features in both examples while Lasso failed Example~\ref{enu: small block}. Probabilities using EGL-R(SA) were lower for irrelevant features, but higher for informative features than EGL-R(S).}
\vskip -0.2in
\label{fig: select prob}
\end{figure}

To evaluate the performance of artificial features, we focused on Examples~\ref{enu: small block} and \ref{enu: er graph}, where Lasso performed similarly or better, with a slightly more challenging task. We set $m = 1000, n = 350, n_S = 50$ and $\sigma^2 = 0.1$. We increased the correlation level $\rho$ for Example~\ref{enu: small block} to 0.99, which is very challenging to Lasso, and $\rho = 0.25$ for Example~\ref{enu: er graph} to ensure positive definite covariance. We set $w_S = 0.2$ and furthermore, we restricted the number of positive and negative weights for informative features to be equal (25 positively weighted and 25 negatively weighted informative features). This is challenging for Lasso since the condition for discovering correct features, the irrepresentable condition~\cite{ir_cond}, is easily violated. For Example~\ref{enu: small block}, we used 10 blocks of 5 features. The connection probability in Example~\ref{enu: er graph} was set to $\frac{4}{49}$ to simulate the connection in Example~\ref{enu: small block}. We used 70 groups in random group allocation. Artificial features and weights were generated such that $X'_{ij} \sim \mathcal{N}(0, 1)$, $\mathbf{w}'_j \sim \mathcal{N}(0, 0.05)$ (Example~\ref{enu: small block}) or $\mathbf{w}'_j \sim \mathcal{N}(0, 0.1)$ (Example~\ref{enu: er graph}). All experiments were performed with stability selection of 50 iterations on 100 randomly generated datasets. Apart from F measure, we also report the average selection probabilities for informative and irrelevant features in Table~\ref{tab: prob}. Selection probabilities are shown in Figure~\ref{fig: select prob}. Lasso performed well on Example~\ref{enu: er graph}, perhaps because of the low correlation ($\rho = 0.25$). However, Lasso failed to select a large proportion of informative features in Example~\ref{enu: small block}. Exclusive group Lasso performed well on both examples. The introduction of artificial features led to lower selection probabilities (more ``compressed'' points as shown in Figure~\ref{fig: select prob}) for irrelevant features but higher probabilities for informative features. Therefore, EGL-R(SA) is promising in more difficult cases.

\begin{table}[tb!]
\centering
\small
\caption{F measure (F) and mean selection probabilities of informative (I) and irrelevant (IR) features using Lasso, EGL-F(S), EGL-R(S), EGL-R(SA) (random groups with stability selection and artificial features). }
\label{tab: prob}
\vskip 0.15in
\begin{center}
\begin{small}
\begin{sc}
\begin{tabular}{lcccccc}
\toprule
 & \multicolumn{3}{c}{Example 3}  & \multicolumn{3}{c}{Example 4}   \\
 \cmidrule(lr){2-4} \cmidrule(lr){5-7}
 & F & I & IR & F & I & IR \\
\midrule
Lasso(S) & 0.79 & 0.61  & 0.01  & 0.99 & 1.00  & 0.13  \\
EGL-F(S) & 1.00 & 1.00  & 0.01  & 1.00 & 1.00  & 0.03  \\
EGL-R(S) & 0.99 & 0.81  & 0.12  & 0.93 & 1.00  & 0.53  \\
EGL-R(SA) & 1.00 & 0.82  & 0.10  & 0.94 & 1.00  & 0.46  \\
\bottomrule
\end{tabular}
\end{sc}
\end{small}
\end{center}
\vskip -0.1in
\end{table}


\section{Real-world application}
\label{sec: cmv}

For real-world application, let us consider the datasets introduced in Section~\ref{sec: lasso}. We generated artificial features for CMV+ and CMV- samples separately such that $\mathbf{x}_{(+)} \sim \mathcal{N}(0.1, 0.5), \mathbf{x}_{(-)} \sim \mathcal{N}(0, 1)$, where $\mathbf{x}_{(+)}$ and $\mathbf{x}_{(-)}$ denote the artificial features of CMV+ and CMV- samples, respectively. The artificial features were added to 200 random groups, each of which contained a single artificial feature. Since CMV-specific TCRs are likely to be more frequent in the CMV-infected group, we restricted the selected features to those that have positive weights with respect to CMV class. Again, exclusive group Lasso was only used for TCR selection, while the classification task was performed by SVM. Stability selection with 30 iterations was also applied to feature selection. Similar to Section~\ref{sec: lasso}, the proposed method was applied to Cohort 1 for training and Cohort 2 and 3 for test. Parameters were selected by a 10-fold cross-validation. Feature selection results and classification accuracy are reported in Table~\ref{tab: cmv result}. For feature selection, EGL-R(SA) selects 47 TCRs, all of which are over-represented in CMV+ samples. Compared to $\ell_1$-regularized SVM, EGL-R(SA) achieved much higher accuracy with a sparse set of biologically meaningful features. The finding that EGL-R(SA) can select a very small set of features (47 from $> 200,000$) which predict CMV status accurately in two independently-generated validation sets suggests the algorithm finds a stable set of informative features. Further biological experiments will be required to validate that these TCRs are indeed specific for CMV.

\begin{table}[tb!]
\centering
\caption{Feature selection and classification accuracy on CMV data. \#TCR: number of selected TCRs, \#CMV+TCR: number of selected TCRs that are over-represented in CMV-infected samples.}
\label{tab: cmv result}
\vskip 0.15in
\begin{center}
\begin{small}
\begin{sc}
\begin{tabular}{llcc}
\toprule
  &  & $\ell_1$-SVM(S) & EGL-R(SA) \\   
  \midrule
  \multicolumn{2}{l}{\#TCR (\#CMV+TCR)} & 1950 (44) & 47 (47)\\
  \midrule
\multirow{ 2}{*}{Cohort 2} & Linear SVM & 0.65  & 0.88 \\  
 & RBF SVM & 0.65  & 0.88 \\   
 \midrule
\multirow{ 2}{*}{Cohort 3} & Linear SVM & 0.51 & 0.89 \\  
 & RBF SVM & 0.63  & 0.89 \\  
\bottomrule
\end{tabular}
\end{sc}
\end{small}
\end{center}
\vskip -0.1in
\end{table}



\section{Conclusion}
\label{sec: concl}

In this paper, we developed fast algorithms to solve exclusive group Lasso, which also allows the possibility to introduce a priori constraints on the values of the weights which may be derived from prior knowledge. Motivated by the challenges of high-dimensional biological datasets, we addressed the problems that informative features are often highly correlated, and neither the number of informative features, nor the underlying correlation structure (i.e. group structure) is usually known. We proposed solutions with stability selection and artificial features for unknown underlying group structure. The exclusive group Lasso, incorporating random group allocation, with stability selection and artificial features, outperformed Lasso and achieved high performance and robust feature selection in a wide range of challenging problems. Future work will be required to further improve model selection and parameter tuning. However, we believe these new feature selection algorithms will prove useful in extracting mechanistic and conceptual understanding from the increasingly complex ``omic'' datasets being generated across the biomedical domain.


\bibliography{example_paper}
\bibliographystyle{icml2020}

\end{document}

%% file: fig/reg.tex
\begin{figure}[htbp]
	\vskip 0.1in
	\centering

	\begin{subfigure}[b]{.33\columnwidth}
	\centering
	\begin{tabular}{|C{0.3pt}| C{0.3pt} |C{0.3pt} | C{0.3pt} | C{0.3pt} | C{0.3pt}|}
	\hline
        \tikzmark{first cell}{}&&&&&\tikzmark{last cell}{}\\
        \hline
        \end{tabular}
        
        \link{first cell}{last cell}{\footnotesize $\ell_1$ norm\\}	

	\caption{Lasso}
	\end{subfigure}%
	~
	\begin{subfigure}[b]{.33\columnwidth}
	\centering
	\begin{tabular}{|C{0.3pt}| C{0.3pt} |C{0.3pt} | C{0.3pt} | C{0.3pt} | C{0.3pt}|}
		\hline
                \tikzmark{cell 1}{\cellcolor{gray!25}}&\tikzmark{cell 2}{\cellcolor{gray!25}}&\tikzmark{cell 3}{}&\tikzmark{cell 4}{}&\tikzmark{cell 5}{\cellcolor{gray!50}}&\tikzmark{cell 6}{\cellcolor{gray!50}}\\
                \hline
	\end{tabular}
	
	\link{cell 1}{cell 2}{\footnotesize $\ell_2$ }
	\link{cell 3}{cell 4}{\footnotesize $\ell_2$ }
	\link{cell 5}{cell 6}{\footnotesize $\ell_2$ }
	\toplink{cell 1}{cell 6}{\footnotesize $\ell_1$ norm}

	\caption{GL}
	\end{subfigure}%
	~
	\begin{subfigure}[b]{.33\columnwidth}
	\centering
	\vspace{39pt}
	\begin{tabular}{|C{0.3pt}| C{0.3pt} |C{0.3pt} | C{0.3pt} | C{0.3pt} | C{0.3pt} |}
		\hline
                \tikzmark{cell 1}{\cellcolor{gray!25}}&\tikzmark{cell 2}{\cellcolor{gray!25}}&\tikzmark{cell 3}{}&\tikzmark{cell 4}{}&\tikzmark{cell 5}{\cellcolor{gray!50}}&\tikzmark{cell 6}{\cellcolor{gray!50}}\\
                \hline
	\end{tabular}
	
	\link{cell 1}{cell 2}{\footnotesize $\ell_1$ }
	\link{cell 3}{cell 4}{\footnotesize $\ell_1$ }
	\link{cell 5}{cell 6}{\footnotesize $\ell_1$ }
	\toplink{cell 1}{cell 6}{\footnotesize $\ell_2$ norm}

	\caption{EGL}
	\end{subfigure}

	\caption{A comparison of the regularization terms of Lasso, group Lasso (GL), and exclusive group Lasso (EGL).}
	\label{fig: compare norm}
	\vskip -0.1in
\end{figure}